# HyperNets and their application to learning spatial transformations

Alexey Potapov[1,2], Oleg Shcherbakov[1,2], Innokentii Zhdanov[1,2], Sergey Rodionov[1,3] and Nikolai Skorobogatko[1,3]

[1] SingularityNET Foundation, Amsterdam, The Netherlands
pas.aicv@gmail.com, astroseger@gmail.com
[2] ITMO University, St. Petersburg, Russia
scherbakovolegdk@yandex.ru, avenger15@yandex.ru
[3] Novamente LLC, Rockville, USA
nicksk@mail.ru

**Abstract.** In this paper we propose a conceptual framework for higher-order artificial neural networks. The idea of higher-order networks arises naturally when a model is required to learn some group of transformations, every element of which is well-approximated by a traditional feedforward network. Thus the group as a whole can be represented as a hyper network. One of typical examples of such groups is spatial transformations. We show that the proposed framework, which we call HyperNets, is able to deal with at least two basic spatial transformations of images: rotation and affine transformation. We show that HyperNets are able not only to generalize rotation and affine transformation, but also to compensate the rotation of images bringing them into canonical forms.

**Keywords:** Artificial neural networks, higher-order models, affine transformation, rotation compensation, currying neural networks, HyperNets

## 1 Introduction

Generalization properties of different neural networks architectures have been of interest since the invention of these type of models. Theoretical and empirical studies of models' generalization properties remain relevant till present [1]. In addition, this problem has a very special place in the field of computer vision: it is crucial for a general-purpose computer vision system to learn the invariant representations of sensor inputs [2]. Classic feedforward discriminative architectures even for deep models have been studied decently, and it seems like their generalization properties are quite restricted since such models cannot directly transfer the results of previous learning to very new domains [3]. Moreover, even whilst working in the same domain but with great variability in data, these models still give very poor results. A very instructive example is the inability of multilayered perceptron to effectively recognize rotated versions of handwritten digits while being trained on canonical ones [4]. Convolutional neural networks partially address the problem of invariant features by mak-ing assumptions of locality and shared parameters. However, these assumptions are yet not enough to force different types of ConvNets learn to distinguish between rotated digits without training on rotated examples [5,6]. Recently, new models named capsule networks have been proposed [7], which are aimed to treat the invariance problem in a very specific way. Capsules are intended to store additional pieces of information in a basic neuron structure that could result in learning of non-trivial spatial relationships between the elements of sensor input on different levels of abstraction. However, training methods for CapsNets are still not as efficient as traditional version of gradient descent due to intensive process of dynamic routing.

It is also interesting that generalization properties of traditional models, which have been trained to reconstruct the original or canonical representations of modified inputs in autoencoder style, are also weak, especially if data domain has changed [8,9].

Nonetheless, one of the most successful techniques to address the problem of geometric transform compensation for input images is the usage of spatial transforming layers [10]. Usually such layer consists of three main parts: localization net, grid generator and sampler. Such architecture allows for explicit spatial manipulation with data within a network for a wide family of parameterized spatial transformations. It has been shown that models with spatial transforming layers generally have increased classification accuracy. However, the concept of ST-layer has several drawbacks: the necessity to choose only differentiable sampling kernels (*e.g.*, the bilinear), explicit representation of parameterized family of transformations, dependency on grid generation and some domain specificity.

As an alternative, we present a technique, which we call hyper-neural networks or simply HyperNets. This is a method for manipulating model parameters by another model. Herewith, one deep neural network can represent both models. Let us consider the HyperNets in more detail.

## 2 Main idea

Consider a network that accepts an image as input and produces its transformed version. The network with dense connections between input and output can easily learn to apply any (but fixed) spatial transformation to an arbitrary image. E.g. it can learn to rotate an image by 45º, or flip an image vertically, etc., but the weights of connections will be different for each individual transformation.

Imagine we want the network to learn how to rotate an image by arbitrary angle provided as an input without hard-coding a special (non-neural) procedure for spatially transforming images. If we add traditional neurons accepting the parameters of transformation as input in addition to image, the network will just mix the image content with these parameters. Even making the network deeper and appending its latent code with the transformation parameters does not help the network to learn how to transform images independently of their content as we shall see later.

Thus, if we have an input image $\mathbf{x} \in X$ and transform parameters $\varphi \in \Phi$, it is convenient to represent transformation process as mapping $X \times \Phi \to X$:

$$\mathbf{x}' = f(\mathbf{x}|\boldsymbol{\varphi}), \tag{1}$$

where **x**′ denotes the transformed image. Given a labeled training set, a traditional model tries to learn an approximation $g$ to the function $f$:

$$f(\mathbf{x}|\boldsymbol{\varphi}) \approx g_\theta(\mathbf{x}, \boldsymbol{\varphi}), \tag{2}$$

where $\theta$ denotes adjustable parameters of the model. In such approach $\varphi$ is usually treated as an additional vector of input values that could be connected to an arbitrary layer of the model presented by a deep network (Fig. 1). As we shall see later in this case the model will not have enough generalization properties.

Instead we can do some form of currying for the function $g$. As a result we obtain a new function **curry**($g$):

$$\mathbf{curry}(g) = h_\omega(\boldsymbol{\varphi}) = (\boldsymbol{\varphi} \mapsto g_\theta(\mathbf{x}, \boldsymbol{\varphi})), \tag{3}$$

which should also have trainable parameters $\omega$. So now we can directly search for the mapping $h: \Phi \to (X \to X)$.

Since it is easy for networks to learn how to transform images by an individual transformation, but their trained weights depend on the parameters of this transformation, it seems quite natural to introduce control neurons, which take these parameters as input and modulate the connection weights of the controlled network through higher-order connections.

The core idea behind HyperNets is representation of neural networks as higher-order functions, which implies a very special network architecture where function $h_\omega(\varphi)$ is a neural network too (Fig. 2). This means that parameters $\theta$ of the network $g$ are described as outputs of the network $h_\omega(\varphi)$: $\theta(\varphi) = h_\omega(\varphi)$. Thus, we try to approximate the target function $f$ by the following model:

$$f(\mathbf{x}|\varphi) \approx g_{h_\omega(\boldsymbol{\varphi})}(\mathbf{x}). \tag{4}$$

In case of complex models $g$, especially some deep ones, not all of the parameters $\theta$ could depend on transformation parameters $\varphi$. Hence, we can rewrite our expression for a higher-order model using a slightly redundant notation:

$$f(\mathbf{x}|\boldsymbol{\varphi}) \approx g_{h_\omega(\boldsymbol{\varphi}),\theta'}(\mathbf{x}), \tag{5}$$

where $\theta'$ denotes the parameters of the model $g$ that are not affected by higher-order terms, i.e. $\theta = \{h_\omega(\boldsymbol{\varphi}), \theta'\}$. Again, having the training set $D$ comprised by $m$ pairs of canonical and transformed images along with respective transformation parameters: $D = \{\mathbf{x_i}, \mathbf{x'_i}, \boldsymbol{\varphi}_i\}_{i=1}^m$, the goal of the model is to learn the transformation concept $h_\omega(\boldsymbol{\varphi})$ and its properties by minimizing some error/loss function and, thus, finding optimal values for $\theta'$ and $\omega$:

$$\omega^*, \theta'^* = \underset{\omega,\theta'}{\mathrm{argmin}}(\textstyle\sum_{i=1}^m L(\mathbf{x'_i}, \mathbf{z}_i = g_{h_\omega(\boldsymbol{\varphi}_i),\theta'}(\mathbf{x}_i))), \tag{6}$$

where $L(\mathbf{x},\mathbf{z})$ is the corresponding loss function between the model's output and the target image.

The name 'hyper network' comes from the analogy between the higher-order functions represented by neural nets and hypergraphs, which could be considered as an extension of traditional computational graph approach.

In this work we have considered relatively simple HyperNet architectures. The interaction between parameters of the higher-order part and the 'core' part of the model presented in Fig. 2 can be described as follows:

$$\mathbf{W}_{hi-ord} = \text{softmax}(\mathbf{a}(\boldsymbol{\varphi})) \qquad (7)$$

$$\mathbf{z} = (\mathbf{W}_{hi-ord} \otimes \mathbf{W})\mathbf{x}, \qquad (8)$$

where **a** is an activation of the last layer of the higher-order part of the network, **W** denotes the parameters (weights) of the 'core' part of the model and ⊗ denotes element-wise product between matrices.

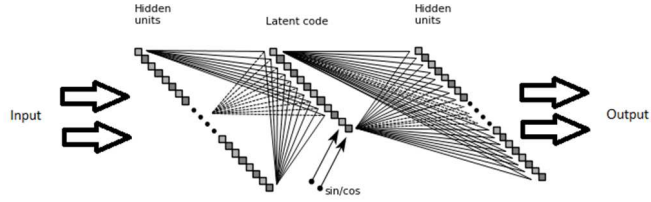

**Fig. 1.** Regular autoencoder with control (sin and cos of desired angle)

It can be seen that the model, whilst being structurally complex, remains differentiable, which allows to directly apply standard optimization techniques under various computational graph frameworks and to simultaneously train both higher-order and 'core' parts of the network. It is also worth saying that transformation parameters $\varphi$ could be represented in numerous ways. For example, in case of planar image rotation $\varphi$ might be parameterized by one angle $\alpha$ with activation constrains for the next hidden layer of the network or by two values representing $\sin(\alpha)$ and $\cos(\alpha)$. So the input to higher-order part of the network will be the transformation parameter, and the input for the 'core' part of the network will be the image. For example, if you want to train a model to rotate an image, you may use cos and sin of an angle as parameters and the unrotated image as an input. The rotated image will be the desired output. Thus, the model is trained in supervised style.

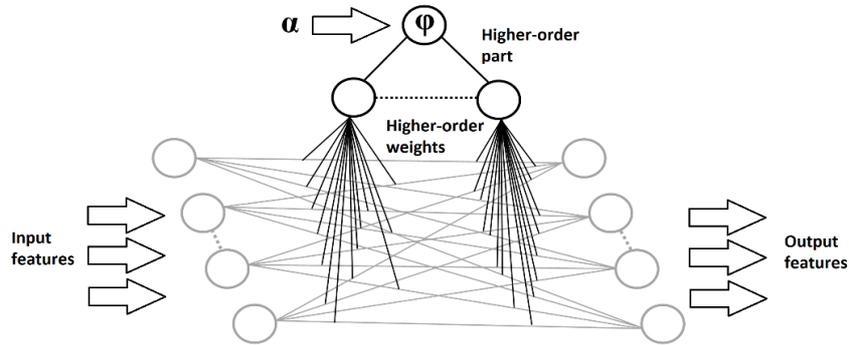

**Fig. 2.** Simple higher-order model architecture

## 3 Experiments and results

We have performed several types of experiments using a developed HyperNet. Also we have tried many architecture modifications, such as adding more dense layers to 'core' network, using convolutional layers, etc.

### 3.1 Rotation experiment

The first experiment is a simple rotation generalization. Previously, we have discussed what will be the input and output in this scenario. In this experiment we have used simple HyperNet, deep HyperNet, deep convolutional autoencoder (AE). Besides of different kind of models, we have tried to learn angles in two different ways: using all angles [0, 360] during training and testing process, and discrete angles (0, 45, 90, 135 … 360 degrees) during training, but all angles while testing (interpolation experiment). Below we presented only results for discrete angles learning for HyperNet and all angles learning for AE. Also, both experiments (with AE and HyperNet) included some extrapolation part for 4 and 9 digits. For these two digits the angles were taken not from [0, 360], but from [0, 90] range of degrees while training. The results of these experiments are shown in Fig. 3.

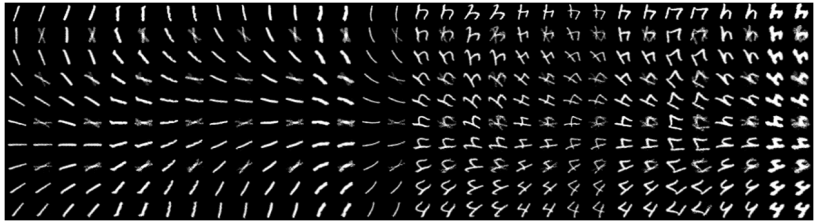

**Fig. 3.** Test results of simple HyperNet trained on discrete angles. Digits 1 and 4

In the figure above, the odd columns contain groundtruth rotated images, the even columns contain images rotated by a simple HyperNet. During testing phase only digits that had not been present in training set were used. Again, as an input to higher-order part of the network sin and cos of the desired angle were used. Moreover, we have tested our model, previously trained solely on digits, to rotate letters. The results are shown in Fig. 4 (left). As can be seen, after adding the higher-order weights, a simple model with just one input and output could be trained to generalize rotation. However, there are some artifacts present on the images, especially if you take a look at digit 1. Hence, a simple model was unable to interpolate rotation on discrete angles learning.

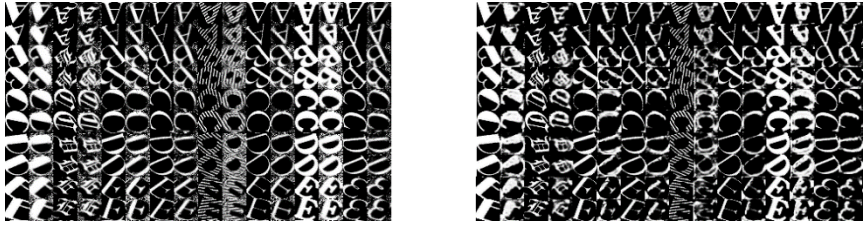

**Fig. 4.** Test results of simple (left) and deep convolutional (right) HyperNet applied to letters. Letters have not been used for training

A slightly deeper model with two convolutional, two dense and two deconvolutional layers (higher-order weights are applied to weights between two dense layers) could return better results, as you can see in Fig. 5. These results were obtained from training on discrete angles and it can be seen that deep higher-order network could already interpolate rotation. At the right part of Fig. 4 the results of testing by deep HyperNet on letters are presented.

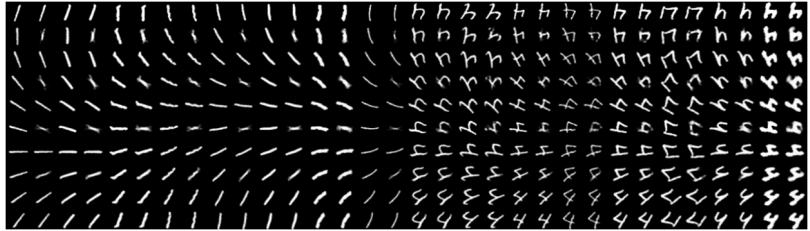

**Fig. 5.** Test results of deep HyperNet trained on discrete angles. Digits 1 and 4

We have also compared reconstruction loss between HyperNet and baseline convolutional autoencoder (AE), which had been designed to learn the rotation transform using information about sin and cos of desired angle added to latent code (see Fig. 7). Ten graphs mean ten digits. Also, in Fig. 6 the results of AE rotation generalization are presented. It is worth saying though, that AE was trained on all angles, not only discrete. It also has to be mentioned, that HyperNet was able to extrapolate rotation representation for 4 and 9 while regular AE with control could not do this.

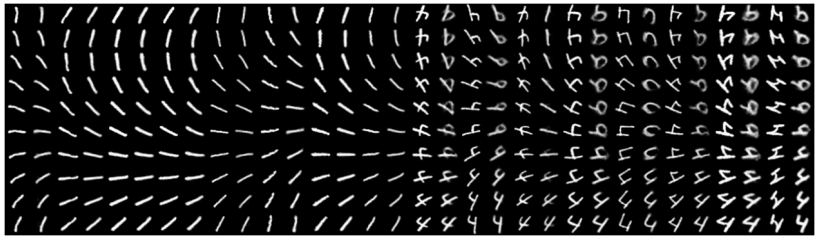

**Fig. 6.** Test results of AE model trained on all angles. Digits 1 and 4

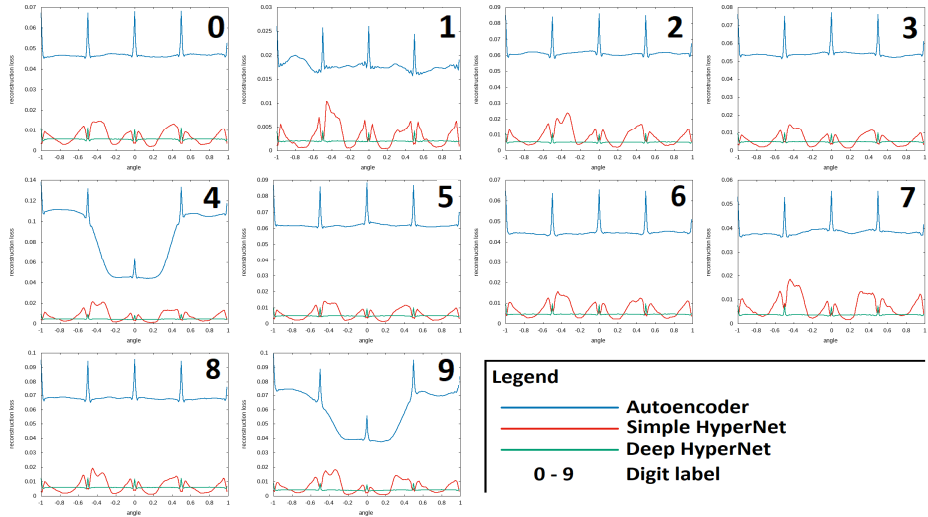

**Fig. 7.** Reconstruction loss comparison of HyperNet models with baseline autoencoder

### 3.2 Affine transformation experiment

Of course, rotation generalization is not such an interesting task. Affine transformation generalization is more challenging. In this case, we have used six affine transformation parameters as a higher-order input, and the canonical image as an input for the core part of the network. The transformed image was used as the desired output. It is worth saying though, that affine transformation parameters were limited in their range to ensure that the digit is still present on the 28×28 image and is recognizable for human. In Fig. 8 you can see the results. In this experiment we are presenting only deep model (with convolutional layers), since simple model shows worse results (though still decent).

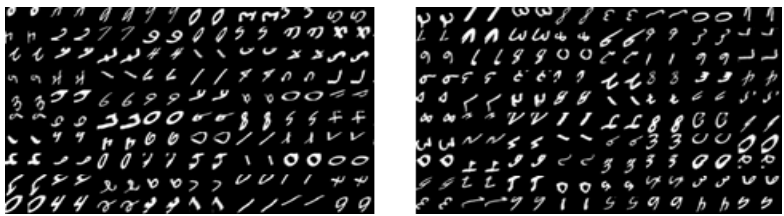

**Fig. 8.** Results of deep HyperNet on the training set (left) and test set (right)

As you can see, the convolutional HyperNet was able to learn almost random affine transformation and to apply it to digits that had not been contained in the training set. Though a simple model still can generalize affine transformation thanks to the higher-order part network, deeper network shows much smoother results. The ability of the AE model to learn affine transformation was also tested and the results are shown in Fig. 9.

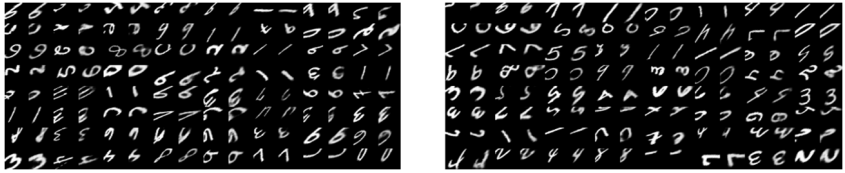

**Fig. 9.** Results of AE model for the training set (left) and test set (right)

The AE model shows decent results, however some blurring and artifacts are present at these pictures. The comparison between AE and HyperNet for the affine transformation generalization experiment is presented in Table 1.

**Table 1.** Comparison between AE and higher-order model

|  | Simple HyperNet | Deep HyperNet | Autoencoder |
|---|---|---|---|
| Reconstruction loss | 0.049395 | 0.0138223 | 0.0606265 |

### 3.3 Rotation compensation experiment

In previous experiments we have tried to learn a model to transform or to simply rotate an input image using control parameters and higher-order architecture. In this last part we were interested in compensating rotation without any knowledge of the angle. This means that control parameters in this scenario will be not sin and cos, since this would be an inverse problem, not so interesting and challenging. But what could be used as an input to the HyperNet then? We have tried to use the rotated image as an input to the core network AND as an input to the higher-order network. And, of course, the canonical image as the desired output. The idea is that the higher-order part of network could possibly extract parameters of the transformation from the rotated image by itself. In this case the dynamics of the higher-order part of network can be described as follows:

$$\mathbf{W}_{hi-ord} = \text{softmax}(\mathbf{a}(\mathbf{x})). \tag{9}$$

But we had to slightly deepen the higher-order part of the network to ensure that it could do such a thing. So, in this experiment, the higher-order part of the network consists of two convolutional layers and one dense layer. Let us see some results in Fig. 10. Only the results of convolutional HyperNet are shown again since it has performed better in previous experiments.

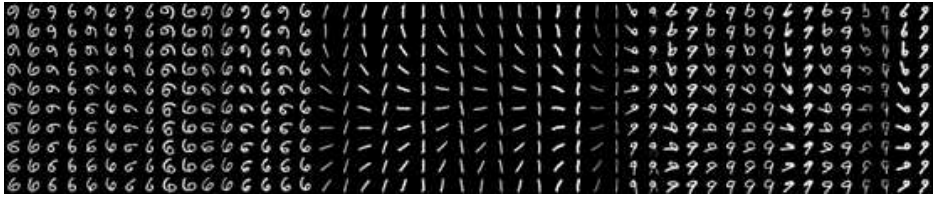

**Fig. 10.** Results of rotation compensation using deep HyperNet. Test set. Digits 6, 1 and 9

In the figures above, the odd columns are the rotated input images and the even columns are the canonical images, which were received from the network. Most interesting results are 6 and 9 digits, since when rotated 180 degrees, 6 actually becomes 9. So, the one could expect that 6 and 9 would be mistaken by the network. However, the HyperNet was able to somehow correctly compensate rotated digits, including 6 and 9. There are some artifacts at the images, but overall the quality is good. Fig. 12 presents the results of AE rotation compensation experiment, and Fig. 11 shows the comparison graphs between these two models. Just to remind, models were trained on digits 4 and 9 that had been rotated only in [0, 90] range of degrees. That explains the difference in graphs for these two digits for the AE model.

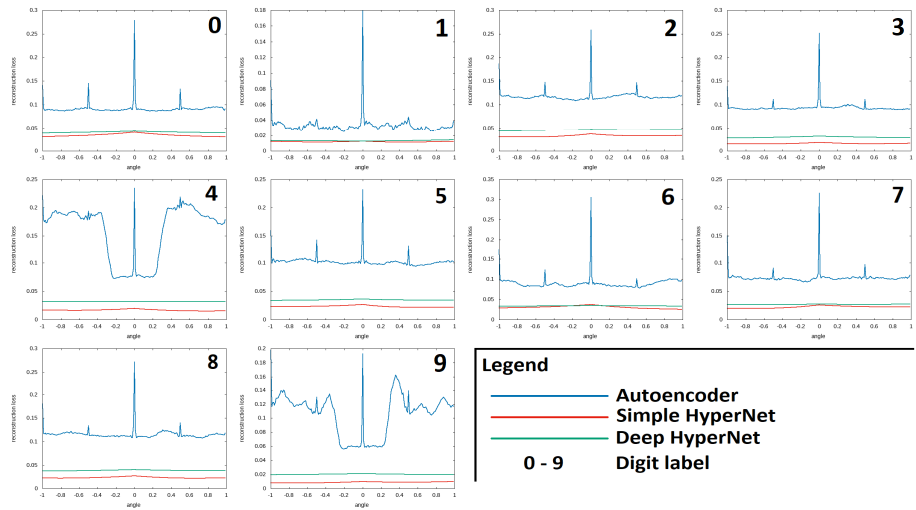

**Fig. 11.** Comparison of HyperNet and autoencoder applied for rotation compensation

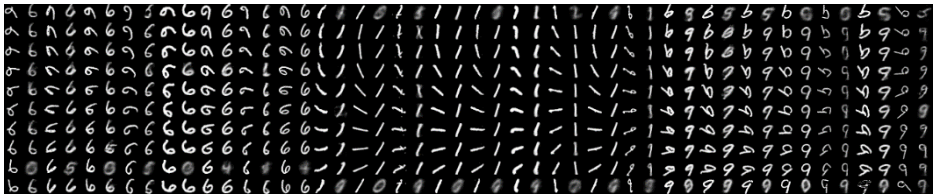

**Fig. 12.** Rotation compensation results using AE model. Test set. Digits 6, 1 and 9

## 4      Conclusion

In this article we have proposed a new approach to artificial neural networks based on generating networks' parameters by higher-order modules that constitute other networks themselves. In other words, the output of the higher-order part acts as a weight matrix for the core part of the network. It has been shown that even a simple Hyper-Net with just one input layer and one output layer in its core part can generalize rotation and affine transformation. The addition of convolution layers allows to receive smoother results. Moreover, deep HyperNet allows to compensate rotation without any information about the angle. In future work it is possible to use such approach to compensate other types of transformations or to extrapolate such approach on generative models.

Our code is availiable on github https://github.com/singnet/semantic-vision/tree/master/experiments/invariance/hypernets